\documentclass[letterpaper, 10 pt, conference]{ieeeconf}  

\IEEEoverridecommandlockouts                              

\usepackage{graphics} 
\usepackage{epsfig} 
\usepackage{mathptmx} 
\usepackage{times} 
\usepackage{cite}
\usepackage{amsmath,amssymb,amsfonts}
\usepackage{textcomp}
\usepackage{xcolor}
\usepackage{booktabs}
\usepackage{algorithm}
\usepackage{algorithmic}
\usepackage{setspace}
\usepackage{ragged2e}
\usepackage{rotating}
\usepackage{subfigure}
\usepackage{multicol}
\usepackage{multirow}
\usepackage{ccaption}
\usepackage{epstopdf}
\usepackage{flushend}
\usepackage{url}
\usepackage{bm}
\usepackage{threeparttable}
\usepackage{array}
\usepackage{makecell}
\usepackage{colortbl}
\usepackage{color}

\let\labelindent\relax
\usepackage{enumitem}

\definecolor{mycyan}{cmyk}{.15,0,0,0}

\title{\LARGE \bf
GOOD: General Optimization-based Fusion for 3D Object Detection via LiDAR-Camera Object Candidates
}

\author{Bingqi Shen$^{1}$, Shuwei Dai$^{2}$, Yuyin Chen$^{2}$, Rong Xiong$^{1}$, Yue Wang$^{1}$, and Yanmei Jiao$^{3}$ 
\thanks{$^{1}$Bingqi Shen, Rong Xiong and Yue Wang$^{*}$ are with the State Key Laboratory of Industrial Control Technology and Institute of Cyber-Systems and Control, Zhejiang University, Hangzhou, China. $^{2}$Shuwei Dai and Yuyin Chen are with Hangzhou Iplus Technology Co., Ltd, Hangzhou, China. Yanmei Jiao is with the School of Information Science and Engineering, Hangzhou Normal University, Hangzhou 311121, China.}%
\thanks{Corresponding author,
        {\tt\small ymjiao@hznu.edu.cn}, Co-corresponding author,
        {\tt\small wangyue@iipc.zju.edu.cn}}%
}

\begin{document}

\maketitle
\thispagestyle{empty}
\pagestyle{empty}

\begin{abstract}

3D object detection serves as the core basis of the perception tasks in autonomous driving. Recent years have seen the rapid progress of multi-modal fusion strategies for more robust and accurate 3D object detection. However, current researches for robust fusion are all learning-based frameworks, which demand a large amount of training data and are inconvenient to implement in new scenes. In this paper, we propose GOOD, a general optimization-based fusion framework that can achieve satisfying detection without training additional models and is available for any combinations of 2D and 3D detectors to improve the accuracy and robustness of 3D detection. First we apply the mutual-sided nearest-neighbor probability model to achieve the 3D-2D data association. Then we design an optimization pipeline that can optimize different kinds of instances separately based on the matching result. Apart from this, the 3D MOT method is also introduced to enhance the performance aided by previous frames. To the best of our knowledge, this is the first optimization-based late fusion framework for multi-modal 3D object detection which can be served as a baseline for subsequent research. Experiments on both nuScenes and KITTI datasets are carried out and the results show that GOOD outperforms by 9.1\% on mAP score compared with PointPillars and achieves competitive results with the learning-based late fusion CLOCs.

\end{abstract}

\section{Introduction}
    It is widely acknowledged that perception systems are the core component of autonomous driving while 3D object detection plays an important role in perception tasks, which is the prerequisite of path planning, motion prediction, collision avoidance, high-definition map generation, etc. According to the input from different modalities, 3D object detection can be divided into image-based \cite{b9} \cite{b10} \cite{b11}, LiDAR-based, and multi-modal-based methods. Considering that LiDAR-based methods are limited to sparse, irregular, and texture-less point cloud while image-based methods suffer from the absence of depth information although they can provide abundant semantic features at very high resolution, many endeavors have been made on multi-modal-based methods to fuse the information from the two complementary sensors for more accurate 3D object detection.

    \begin{figure}[t]
        \centering
        \includegraphics[width=0.5\textwidth]{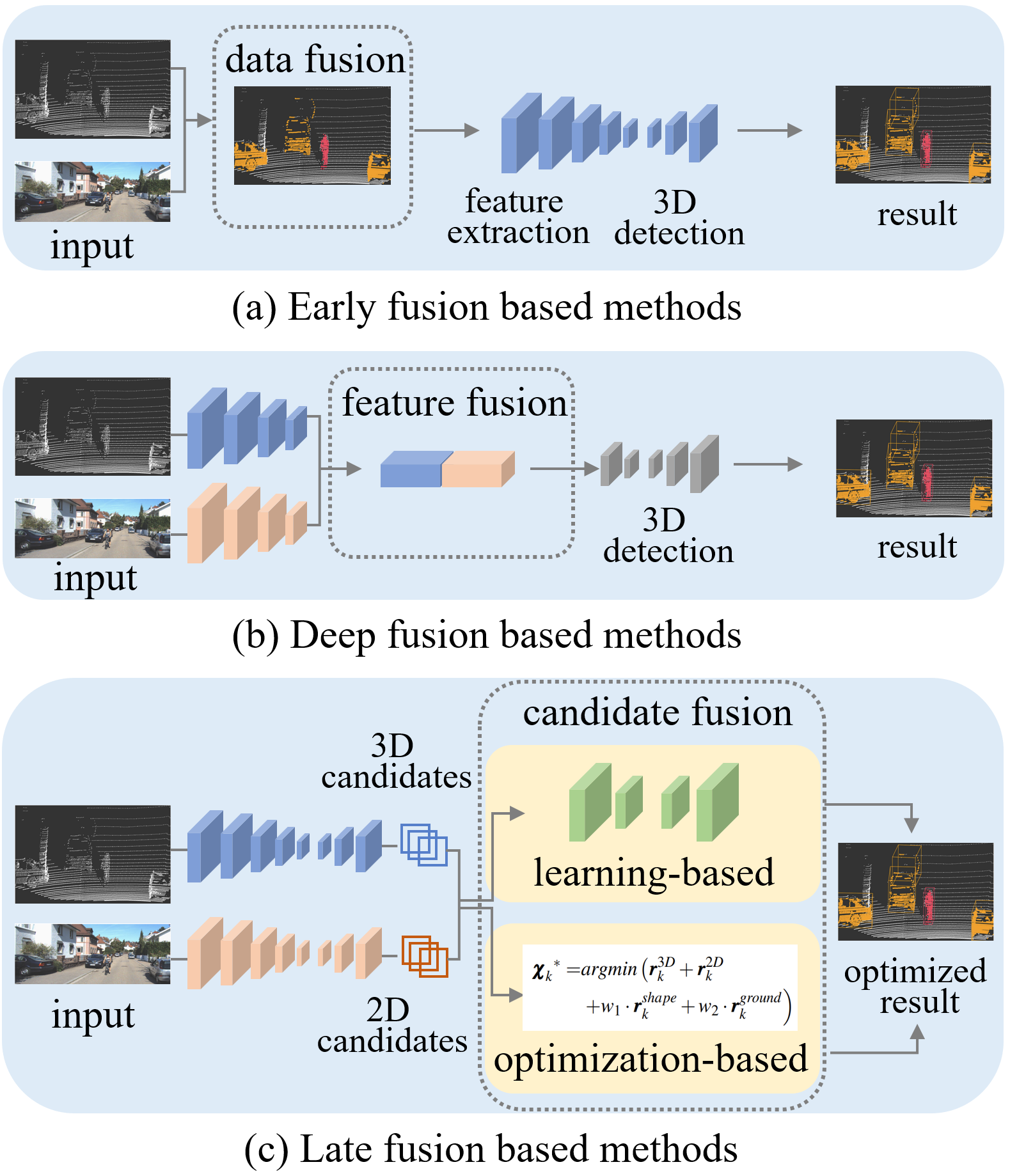} 
    
        \caption{An illustration of different types of LiDAR-camera fusion for 3D object detection methods. The proposed GOOD is a general optimization-based late fusion method, which (1) is distinguished from early or deep fusion that demand pixel-wise correspondences and high-precision calibration (2) and do not require designing additional fusion network for different 3D/2D detectors like learning-based late fusion (3) as well as is free from training stage.}
        \vspace{-0.4cm}
    \end{figure}
    
    Depending on the stage where observations from different modalities are fused, current LiDAR-camera fusion solutions can be categorized into early fusion, deep fusion, and late fusion, which correspond to the fusion at the input, intermediate, and output stages respectively as shown in Fig. 1. Although early fusion \cite{b12} and deep fusion \cite{b20} strategies can make full use of the information from multiple modalities and achieve remarkable performance in terms of accuracy, both of them demand high accuracy on view alignments and synchronization. When it comes to autonomous driving in reality, chances are that more noise will be introduced on account of sensor calibration and synchronization, which are the important factors that lead to low robustness and worse performance compared with pure LiDAR-based methods. Besides, they generally introduce a relatively complicated network structure that improves the complexity of the whole system.

    Therefore, some late fusion strategies \cite{b14} \cite{b15} have been investigated in recent years to improve accuracy while maintaining higher robustness in practical applications. Nevertheless, both of them still design a network to learn the association of 3D-2D boxes and obtain the final results, which demands large-scale training data and is hard to be implemented in new scenes. In addition, the network structure needs redesigning once the target class of detection changes. Towards this goal, we propose an optimization-based late fusion framework fusing the candidates from LiDAR and camera to achieve a better 3D detection effect. First, we apply a mutual-sided nearest-neighbor probability model to match the detection information from LiDAR and camera. Then we use an optimization pipeline, including $refine$, $add$, and $keep$ $or$ $delete$ modules, to optimize different kinds of instances after matching separately. Besides, we introduce the 3D Multiple Object Tracking (MOT) module that can further optimize the 3D detection bounding boxes by previous frames. Experiments on both nuScenes and KITTI dataset are carried out and the results show that GOOD outperforms by 9.1\%, 6.4\%, and 3.9\% in the mAP score and 6.2\%, 5.0\%, and 4.2\% in the NDS score compared with corresponding LiDAR-based baselines: PointPillars, SECOND, and CenterPoint respectively and achieves competitive results with the learning-based late fusion CLOCs. To the best of our knowledge, this is the first work on learning-free late fusion for 3D object detection without a large amount of training data and retraining models facing new scenes. Furthermore, this fusion method can be applied on the lightweight hardware platform without GPUs since low computing resources are required while the 3D/2D detectors can be deployed on other high performance computing platforms in application in virtue of lower data transmission. In summary, the main contributions of this paper are that:

    1) A general optimization-based fusion for 3D object detection (GOOD) framework is designed which can leverage any 3D/2D detectors without any change. 
    
    2) An optimization pipeline, including $refine$, $add$, and $keep$ $or$ $delete$ modules, is proposed and the mutual-sided nearest-neighbor probability model is applied.
    
    3) Extensive evaluations are performed to validate the strength of the system as well as the effectiveness of each component in the framework.
    
    The rest of this paper is organized as follows: Section II discusses related work on 3D object detection, and Section III elaborates on our proposed framework. For validation, we conduct relevant experiments and the results are presented in Section IV. Ultimately, the whole work is concluded in Section V.

\section{Related Work}
    \subsection{LiDAR-based 3D detection}
    Depending on the representation learning strategies, the existing works fall into point-based, voxel-based, and point-voxel-based. Point-based methods propose a diverse architecture to detect 3D objects directly from the raw point cloud. PointRCNN \cite{b17} leverages PointNet-like block \cite{b24} to generate 3D proposals in a bottom-up manner using point cloud segmentation and these proposals are refined in the second stage to generate the final detection boxes. This kind of method retains all information from the point cloud to achieve high accuracy at the cost of efficiency. In contrast, voxel-based methods try to voxelize the irregular point cloud to pillars \cite{b25}, voxels \cite{b16} or frustums \cite{b26}, and then transform them into bird's-eye view (BEV) 2D representation, where convolution neural networks (CNNs) can be applied to detect 3D objects. In this way, they can be easily amenable to efficient hardware implementations. Point-voxel-based methods such as \cite{b18} adopt \cite{b16} as the first-stage detector, and the RoI-grid pooling operator is proposed for the second-stage refinement. They can benefit from the fine-grained 3D shape and structure information obtained from points and the computational efficiency brought by voxels.

    \subsection{Multi-modal-based 3D detection}
    As mentioned above, this kind of work includes three categories. Early fusion strategies aim to incorporate the knowledge from images into the point cloud at the stage before they are fed into a 3D detection network. PointPainting \cite{b12} leverages image-based semantic segmentation to augment point cloud while \cite{b19} proposes a novel geometric agreement search. Deep fusion strategies try to fuse image and LiDAR features at the proposal generation stage. MV3D \cite{b20} projects the point cloud into BEV to form a BEV feature map. Then a 2D convolutional neural network (CNN) will be adopted to extract features from these BEV images as well as the front camera image for 3D bounding box generation. AVOD \cite{b13} further extends the fusion strategy to the first stage to enrich more informative semantics for proposal generation. Late fusion strategies fuse the outputs of different modalities like 3D/2D bounding boxes at the decision stage, which not only avoid the issue of alignments and synchronization but also are more efficient compared with other approaches. CLOCs \cite{b14} introduces a sparse tensor that contains paired 3D-2D boxes and learns the final object confidence scores from the tensor while \cite{b15} improves by introducing a lightweight 3D detector-cued image detector.

    \begin{figure*}[t]
        \centering
        \includegraphics[width=0.9\textwidth]{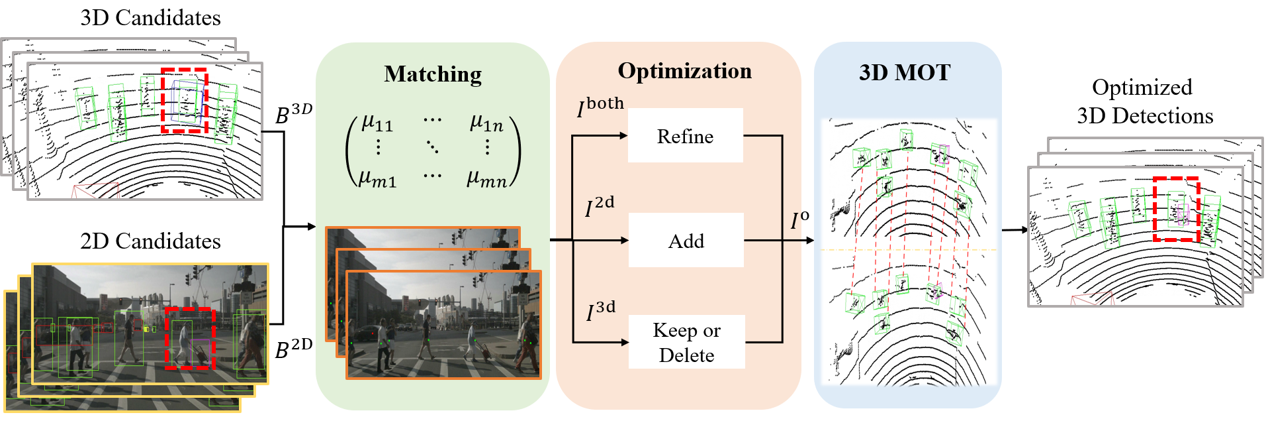} 
    
        \caption{An overview of our proposed framework: At the input, we obtain the 3D/2D candidates from LiDAR-based 3D object detector and image-based 2D detector. Next, we match the 3D candidates with the 2D candidates and all the candidates are divided into three kinds of instances in result and the optimization will be conducted where they are handled separately. Then we pass the optimized instances through the 3D MOT procedure which can further optimize the 3D boxes utilizing prediction from the previous frames to obtain the final 3D detections.}
        \vspace{-0.2cm}
    \end{figure*}
    

\section{Methodology}

    In this section, we elaborate on our proposed method as shown in Fig. 2. There are three main parts in GOOD including matching, optimization, and 3D MOT. The inputs of the whole system are 3D/2D candidates from a single or sequence of point cloud and images given by the 3D/2D detector. And we consider the output of a set of optimized 3D detection bounding boxes for each frame.

    \subsection{Assumptions and notations}\label{AA}
        We assume that the intrinsic parameters of each camera are known. The extrinsic parameters between LiDAR and cameras are calibrated, and they are time synchronized. 
    
        Let us denote the output of the 3D object detector at timestamp $t$ as $\bm{B}_{t}^{3D}$, which refers to the 3D object-oriented bounding boxes in LiDAR coordinate and confident scores. Supposing there are $m$ 3D detection bounding boxes at timestamp $t$, it can be defined as follows:
        
        \begin{equation}
            \begin{split}
                \bm{B}_{t}^{3D}&=\left \{ \bm{b}_{1}^{3D}, \bm{b}_{2}^{3D}, ..., \bm{b}_{i}^{3D}, ..., \bm{b}_{m}^{3D} \right \}   \\
                \bm{b}_{i}^{3D}&=\left \{ [x_{i}^{3D}, y_{i}^{3D}, z_{i}^{3D}, l_{i}^{3D}, w_{i}^{3D}, h_{i}^{3D}, \theta_{i}^{3D}], s_{i}^{3D} \right \}
            \end{split}
        \end{equation}
    
        \noindent where $\bm{b}_{i}^{3D}$ is the $i^{th}$ detection while $[x_{i}^{3D}, y_{i}^{3D}, z_{i}^{3D}, l_{i}^{3D}, w_{i}^{3D}, h_{i}^{3D}, \theta_{i}^{3D}]$ is the 7-digit vector for 3D bounding box containing 3D location $(x,y,z)$, 3D dimension $(l,w,h)$ and rotation (yaw angle, $\theta$) since the objects are located on the ground by default. $s_{i}^{3D}$ is the detection score. In this way, the 2D detections at timestamp $t$ can be denoted as $\bm{B}_{t}^{2D}$ in token of 2D bounding boxes in the image plane, which is defined as:
    
        \begin{equation}
            \begin{split}
                \bm{B}_{t}^{2D}&=\left \{ \bm{b}_{1}^{2D}, \bm{b}_{2}^{2D}, ..., \bm{b}_{j}^{2D}, ..., \bm{b}_{n}^{2D} \right \}   \\
                \bm{b}_{j}^{2D}&=\left \{ u_{j1}, v_{j1}, u_{j2}, v_{j2}, s_{j}^{2D} \right \}
            \end{split}
        \end{equation}

        \noindent where $n$ is the total number of 2D bounding boxes and $\bm{b}_{j}^{2D}$ refers to the $j^{th}$ 2D box. $u_{j1}, v_{j1}$ and $u_{j2}, v_{j2}$ are the pixel coordinates of the top left and bottom right corner points from the bounding box. $s_{j}^{2D}$ is the detection score.
    
    \subsection{Matching}
        Our matching method is called mutual-sided nearest-neighbor probability model develops from \cite{b1}. We first project the center point of the 3D bounding box into the image plane so that we can obtain its Euclidean Distance from the center point of the 2D bounding box. Based on this, we calculate the association probability from 3D to 2D as follows:
    
        \begin{equation}
            {P_{\bm{b}_{i}^{3D}} ( \bm{b}_{j}^{2D} ) = \frac{{dist(\bm{b}_{i}^{3D}, \bm{b}_{j}^{2D})}^{-\alpha}}{\sum \limits_{k=1}^n dist{(\bm{b}_{i}^{3D}, \bm{b}_{k}^{2D})}^{-\alpha}}}
        \end{equation}
    
        \noindent where $dist$ refers to the function of calculating Euclidean Distance. $\alpha$ is a weight coefficient that can be set as 0.5. And we can obtain $P_{\bm{b}_{j}^{2D}} ( \bm{b}_{i}^{3D} )$ in the same way to calculate the correlation confidence of $\left \{ \bm{b}_{i}^{3D}, \bm{b}_{j}^{2D} \right \}$ as follows. 
    
        \begin{equation}
            conf(\left \{ \bm{b}_{i}^{3D}, \bm{b}_{j}^{2D} \right \}) = \left\{
            \begin{aligned}
                \sqrt{P_{i} ( j ) \cdot P_{j} ( i )} && c.i = c.j \\
                0 && c.i \neq c.j
            \end{aligned}
            \right.
        \end{equation}
    
        \noindent where $P_{i} (j)$ and $P_{j} (i)$ represent $P_{\bm{b}_{i}^{3D}} ( \bm{b}_{j}^{2D} )$ and $P_{\bm{b}_{j}^{2D}} ( \bm{b}_{i}^{3D} )$ while $c.i$ and $c.j$ are short for the class of the $i^{th}$ 3D bounding box and the $j^{th}$ 2D bounding box. Thus, the confidence that the $i^{th}$ 3D bounding box is not associated with any 2D bounding box can be calculated:
    
        \begin{equation}
            conf(\left \{ \bm{b}_{i}^{3D} \right \}) = 1 - \sum \limits_{k=1}^n \sqrt{P_{i} ( k ) \cdot P_{k} ( i )}
        \end{equation}
    
        Next, a matching matrix $M$ with $(m+1) \times (n+1)$ dimension can be established, whose element in row $i$ and column $j$, denoted $\mu_{ij}$, is defined as follows:
    
        \begin{equation}
            \mu_{ij} = \left\{
            \begin{aligned}
                conf(\left \{ \bm{b}_{i}^{3D}, \bm{b}_{j}^{2D} \right \}) && 1 \leq i \leq m, 1 \leq j \leq n \\
                conf(\left \{ \bm{b}_{i}^{3D} \right \}) && 1 \leq i \leq m, j = n + 1 \\
                conf(\left \{ \bm{b}_{j}^{2D} \right \}) && i = m + 1, 1 \leq j \leq n \\
                0 && i = m+1, j = n+1
            \end{aligned}
            \right.
        \end{equation}
    
        We consider the $i^{th}$ 3D bounding box is associated with the $j^{th}$ 2D bounding box if $\mu_{ij}$ is the maximum value of both the $i^{th}$ row and the $j^{th}$ column of the matching matrix.

    \subsection{Optimization}
        After the matching stage, all the 3D and 2D candidates will be divided into three categories: matched instances ($I^{both}$), unmatched 3D instances ($I^{3D}$), and unmatched 2D instances ($I^{2D}$). We treat them separately with three different modules to obtain the 3D result with higher quality.

        \begin{enumerate}[leftmargin = 0pt, itemindent = 30pt]
            \item \textbf{Refine:}
            After data association, we build error equations and optimize the 3D detection frame associated with 2D annotation information. On the premise that the default object is on the ground and the rotation only takes into account the angle of yaw, the 3D frame state is defined as a seven-dimensional vector. The optimal estimation of 3D semantic frame is obtained by minimizing the residual of laser 3D detection and image 2D annotation.

            \begin{equation}
                \begin{split}
                   {\bm{\chi}_k}^* = &arg min \left(\bm{r}^{3D}_{k} + \bm{r}^{2D}_{k} \right.\\
                   &\left. + w_1 \cdot \bm{r}^{shape}_{k} + w_2 \cdot \bm{r}^{ground}_{k}\right) \\
                \end{split}
            \end{equation}

            \noindent where $\bm{r}^{3D}_{k}$, $\bm{r}^{2D}_{k}$, $\bm{r}^{shape}_{k}$ and $\bm{r}^{ground}_{k}$ are different kinds of residuals which will be explained as follows. $w_1$ and $w_2$ are weight parameters of two residuals. We set $w_1 = 1.2$, $w_2 = 2.0$ after a manual search on small sample datasets.

            \begin{itemize}[leftmargin = 15pt, parsep=5pt, topsep=5pt]
                \item \emph{IoU residuals $\bm{r}^{3D}_{k}$ and $\bm{r}^{2D}_{k}$}: Ideally, the bounding box after optimization is supposed to coincide with the detection bounding box in 3D space and the image bounding box in 2D plane. Therefore, we define $\bm{r}^{3D}_{k}$ as the residual of LiDAR detection associated with $\bm{b}_k^{3D}$, and $\bm{r}^{2D}_{k}$ as the residual of image detection associated with $\bm{b}_k^{2D}$. Both of them are calculated by 3D/2D IoU (Intersection over Union):

                \vspace{-0.5cm}
                \begin{equation}
                    \begin{split}
                       &\bm{r}^{3D}_{k} = {IoU}^{3D}\left( \bm{\chi}_k, \bm{b}_k^{3D}\right) \\
                       &\bm{r}^{2D}_{k} = {IoU}^{2D}\left( \bm{\chi}_k, \bm{b}_k^{2D}\right)
                    \end{split}
                \end{equation}

                \begin{algorithm}[!h]
                    \caption{Depth and z-coordinate of the reconstructed cuboid adjustment algorithm}
                    \begin{algorithmic}[1]
                        \REQUIRE ~~\\
                        original cuboid's depth $\hat{d}$, height $\hat{h}$, z-coordinate $\hat{z}$,\\
                        original point cloud $\bm{S}$, radius $r$, \\
                        maximum number of iteration $N$ \\
            
                        \ENSURE ~~\\
                        optimized cuboid's depth $\widetilde{d}$, z-coordinate $\widetilde{z}$ \\
                    
                        \STATE Segment ground points $\bm{S}_g$ and non-ground points $\bm{S}_n$ from $\bm{S}$
                        \STATE $i \gets 0$, $d \gets \hat{d}$, $z \gets \hat{z}$
                        \STATE Calculate the number of non-ground points contained by the current cuboid $n_0$
                        \WHILE{$i < N$}
                            \STATE $i \gets i + 1$
                            \STATE $d \gets d + \Delta d$
                            \STATE Calculate the number of non-ground points contained by the current cuboid $n_i$
                        \IF{$n_i < n_{i-1}$}    
                            \STATE $d \gets d - \Delta d$
                            \STATE Filter out $S_g^r$ from $S_g$ with radius $r$ centered on the current cuboid
                            \STATE Calculate the average height $h_g^r$ of $S_g^r$
                            \STATE $z \gets h_g^r + \frac{\hat{h}}{2}, i \gets N + 1$
                        \ENDIF    
                        \ENDWHILE
    
                        \IF{$i = N$} 
                            \STATE $\widetilde{d} \gets \hat{d}$
                            \STATE $\widetilde{z} \gets \hat{z}$  
                        \ELSE
                            \STATE $\widetilde{d} \gets d$
                            \STATE $\widetilde{z} \gets z$
                        \ENDIF 
                    \end{algorithmic}
                \end{algorithm}

                \item \emph{Shape residual $\bm{r}^{shape}_{k}$}: As mentioned in \cite{b2}, the similar 2D bounding box can correspond with different 3D cuboids with different skew ratios (length/width). Thus, a shape residual is added to ensure that the shape of the cuboid after optimization is similar to most objects in the same class:

                \vspace{-0.5cm}
                \begin{equation}
                    \begin{split}
                       &\bm{r}^{shape}_{k} = |\frac{l_k}{w_k} - \mu|
                    \end{split}
                \end{equation}

                where $\mu$ is the prior reference value and it varies in different classes.

                \item \emph{Ground residual $\bm{r}^{ground}_{k}$}: Considering that all objects are located on the ground, it is feasible for us to introduce ground constraint which means that the height of cuboid's bottom should be the same as that of the ground on which it stands:

                \vspace{-0.3cm}
                \begin{equation}
                    \begin{split}
                       &\bm{r}^{ground}_{k} = |z_k - \frac{h_k}{2} - z^{g}|
                    \end{split}
                \end{equation}

                \noindent where $z^{g}$ refers to the height of the ground which is obtained by \cite{b3} in this paper. 

            \end{itemize}

            \item \textbf{Keep or Delete:} 
            For unmatched 3D instances from detection bounding boxes, chances are that it may be detected incorrectly like misdetecting a pole as a pedestrian or a lady with a cart as a cyclist as shown in Fig. 2. In this event, a threshold will be set to filter out those incorrect detections whose confident score $s_i$ is below the threshold while those are above the threshold will be reserved.

            \vspace{-0.3cm}

            \begin{table*}[htbp] 
        \caption{\label{tab:test}Performance comparison of different LiDAR-based 3D object detection baselines and our proposed method on nuScenes trainval set. We report NDS, mAP, mATE, mASE, mAOE, mAVE, and mAAE. $\uparrow$ indicates that higher is better and $\downarrow$ indicates that lower is better.} 
        \centering
        \vspace{0.3cm}
        \setlength{\tabcolsep}{2.0mm}
        \begin{threeparttable}
        \begin{tabular}{cccccccc} 
            \toprule 
            Method & NDS $\uparrow$ & mAP $\uparrow$ & mATE $\downarrow$ & mASE $\downarrow$ & mAOE $\downarrow$ & mAVE $\downarrow$ & mAAE $\downarrow$\\ 
            \midrule 
            PointPillars (baseline) & 0.679 & 0.594 & 0.253 & 0.201 & 0.182 & 0.209 & 0.341  \\
            Ours (PointPillars + YOLOv5) & 0.721 & 0.648 & 0.243 & 0.192 & 0.179 & 0.208 & 0.337\\ 
            \rowcolor{mycyan} \textbf{improvement} & +6.2\% & +9.1\% & -3.9\% & -4.5\% & -1.6\% & -0.5\% & -1.2\%\\
            SECOND (baseline) & 0.723 & 0.661 & 0.223 & 0.190 & 0.139 & 0.187 & 0.340  \\
            Ours (SECOND + YOLOv5) & 0.759 & 0.703 & 0.216 & 0.183 & 0.141 & 0.187 & 0.339 \\ 
            \rowcolor{mycyan} \textbf{improvement} & +5.0\% & +6.4\% & -3.1\% & -3.7\% & +1.4\% & -0.0\% & -0.3\% \\
            CenterPoint (baseline) & 0.768 & 0.760 & 0.195 & 0.192 & 0.255 & 0.143 & 0.333 \\
            Ours (CenterPoint + YOLOv5) & 0.800 & 0.790 & 0.195 & 0.183 & 0.246 & 0.143 & 0.332  \\ 
            \rowcolor{mycyan} \textbf{improvement} & +4.2\% & +3.9\% & -0.0\% & -4.7\% & -3.5\% & -0.0\% & -0.3\%  \\
            \bottomrule 
        \end{tabular} 
        \end{threeparttable}
    \end{table*}

    \begin{table*}[!htbp] 
        \caption{\label{tab:test}Performance comparison of multi-modal-based late fusion CLOCs and our proposed method (without 3D MOT) on KITTI validation set. We report 3D AP of car, pedestrian, and cyclist at three levels and the IoU threshold are 0.7, 0.5, and 0.5 respectively.} 
        \centering
        \vspace{0.3cm}
        \setlength{\tabcolsep}{2.0mm}
            \begin{tabular}{c c c c c c } 
                \toprule
                \multicolumn{3}{c}{\multirow{2}{*}{Method}} & \multicolumn{3}{c}{\makecell[c]{3D AP (\%) \\ Car / Pedestrian / Cyclist}} \\  
                \cmidrule(lr){4-6}
                \multicolumn{3}{c}{}& easy & moderate & hard \\  
                \hline 
                \multicolumn{3}{c}{SECOND} & 80.89 / 64.31 / 87.45 & 76.63 / 58.43 / 73.78 & 74.57 / 55.76 / 71.32 \\   
                \multicolumn{3}{c}{CLOCs (SECOND + YOLOv5, 30 epochs)} & 80.65 / \textbf{66.51} / 88.59 & \textbf{78.43} / 61.31 / \textbf{86.14} & 76.71 / \textbf{60.98} / \textbf{79.15} \\
                \multicolumn{3}{c}{Ours (SECOND + YOLOv5, without 3D MOT)} & \textbf{82.43} / 65.87 / \textbf{90.06} & 77.89 / \textbf{61.45} / 75.56 & \textbf{76.96} / 58.71 / 72.04 \\
                \bottomrule 
            \end{tabular} 
        
    \end{table*}
    
            \vspace{0.3cm}
            
            \item \textbf{Add:}
            Given the fact that 3D missed detection is prone to happen owing to the sparse measurement of LiDAR, there are unmatched 2D instances such as the cart pushed by an old lady in Fig. 2. In that case, we refer to the single image 3D object detection proposed by \cite{b2} to reconstruct the 3D cuboid from the 2D bounding box. However, the depth estimation of the cuboid is often inaccurate under the constraint from the single frame observation or due to the fact that the object locates on the pavement which is above the road, causing inaccuracy of z-coordinate estimation. Therefore, we decide to utilize the information generated by ground segmentation to assist adjusting the depth and z-coordinate estimation. Firstly, we can segment the ground points from the original point cloud and calculate the height of the ground around the original reconstructed cuboid. Considering that the azimuth estimation is relatively accurate so that the camera, the reconstructed cuboid, and the point cloud corresponding to the object (belonging to the non-ground point cloud) usually lie in the same line, we can adjust the depth of the cuboid along this line to make it contain more points and then adjust the z-coordinate of the cuboid to ensure that it locates on the ground, thus obtaining the accurate depth and z-coordinate of cuboid as described in Algorithm 1.

        \end{enumerate}

    \subsection{3D MOT}
    The multi-object tracking module can not only give each object a unique ID and associate the same object in successive frames but also provide a more precise detection result by utilizing sequential observations. In this paper, we choose to take the tracking-by-detection path since we have already got the detection results based on the above. Compared with some learning-based methods \cite{b5} \cite{b6} \cite{b7}, our tracker is an implementation variant of the \cite{b4}, which is simpler and more efficient by using detector boxes for associations and Kalman ﬁlter for state updates.

\section{Experiments}

    For the purpose of ensuring the authenticity and objectivity of the experimental results, we conduct two controlled experiments to testify performance of GOOD compared with LiDAR-based and multi-modal-based method on the open datasets: nuScenes \cite{b8} and KITTI \cite{b29} respectively. 
    
    \textbf{nuScenes:} nuScenes provides a large collection of LiDAR sequences with 2D labels available for each frame. It contains 1000 sequences with 700 for training, 150 for validation, and 150 for testing and 10 classes are annotated for detection evaluation. Each sequence lasts around 20 seconds with a sampling frequency at 10Hz. Apart from the mean of Average Precision (mAP), nuScenes uses NuScenes Detection Score (NDS) as its oﬃcial metric for evaluating the detector. 
        
    \begin{equation}
        \begin{split}
           &NDS = \frac{1}{10} [ 5mAP + \sum \limits_{e\in\epsilon} (1-min(1,e))]
        \end{split}
    \end{equation}

    where $\epsilon = \{mATE, mASE, mAOE, mAAE, mAVE\}$ is the error subset of translation, size, orientation, attribute, and velocity. It is worth mentioning that mAP is calculated by a bird-eye-view center distance of thresholds {0.5m, 1m, 2m, 4m} rather than standard box-overlap.

    \begin{table*}[htbp] 
        \caption{\label{tab:test}Ablation studies on the improvements to PointPillars. mAVE and mAAE are not taken into account due to slight shifts. KoD is short for $keep$ $or$ $delete$.}
        \centering
        \vspace{0.3cm}
        \setlength{\tabcolsep}{1.8mm}
        \begin{threeparttable}
            \begin{tabular}{ccccccccccc} 
                \toprule 
                Method & Baseline & KoD & Refine & Add & 3D MOT & NDS $\uparrow$ & mAP $\uparrow$ & mATE $\downarrow$ & mASE $\downarrow$ & mAOE $\downarrow$ \\ 
                \midrule 
                PointPillars & \checkmark & - & - & - & - &  0.679 & 0.594 & 0.253 & 0.201 & 0.182 \\
                \midrule 
                Ours & \checkmark & \checkmark & - & - & - &  +1.6\% & +1.8\% & -1.1\% & -0.4\% & -0.5\%\\
                Ours & \checkmark & \checkmark & \checkmark & - & - & +3.4\% & +2.8\% & -3.4\% & -3.7\% & -1.5\%\\
                Ours & \checkmark & \checkmark & \checkmark & \checkmark & - & +5.7\% & +8.4\% & -2.9\% & -2.5\% & -1.5\% \\
                Ours & \checkmark & \checkmark & \checkmark & \checkmark & \checkmark & \textbf{+6.2\%} & \textbf{+9.1\%} & \textbf{-3.9\%} & \textbf{-4.5\%} & \textbf{-1.6\%} \\
                \bottomrule 
            \end{tabular} 
        \end{threeparttable}
    \end{table*}
        
     \begin{figure*}[t]
        \centering
        \includegraphics[width=1.0\textwidth]{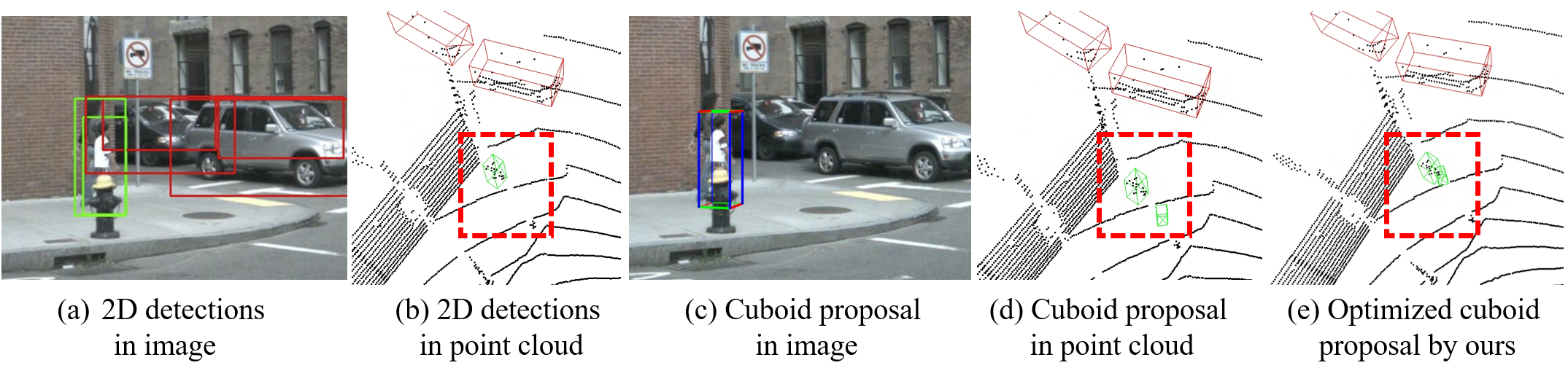} 
    
        \caption{An example of depth and z-coordinate of the reconstructed cuboid adjustment algorithm: A kid and his mother are both detected by the 2D detector in (a) but the kid is missed by the 3D detector in (b). The single image 3D object detection is carried out on unmatched 2D instance in (c) while the depth is not accurate in (d). We utilize the algorithm to adjust its depth in (e). A red dashed rectangle is used in (b) (d) (e) for recognition.}
        \vspace{-0.2cm}
    \end{figure*}
    
    \begin{figure}[t]
        \centering
        \subfigure[2D detections in the image's view]{
            \begin{minipage}[t]{0.93\linewidth}
            \centering
            \includegraphics[height=1.4in,width=\textwidth]{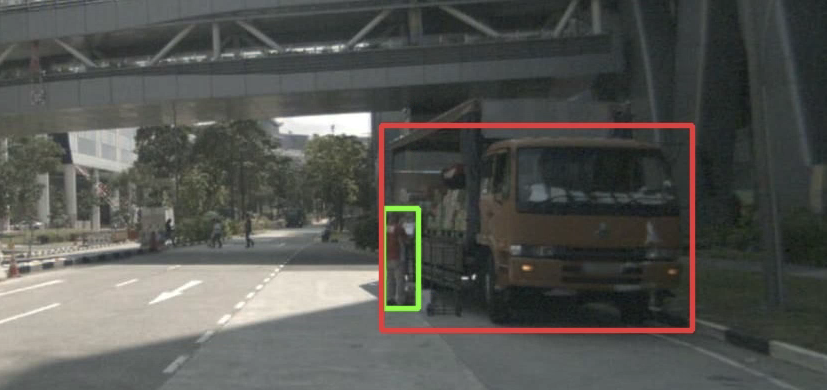} \\
            \end{minipage}
        }

        \subfigure[Original 3D detections in the point-cloud's view]{
            \begin{minipage}[t]{0.45\linewidth}
            \centering
            \includegraphics[height=1.7in,width=\textwidth]{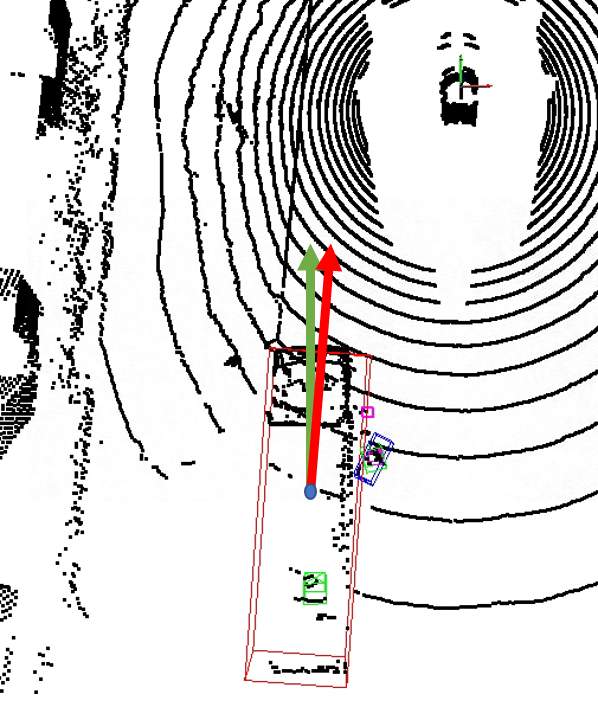} \\
            \end{minipage}
        }
        \subfigure[Refined 3D detections in the point-cloud's view]{
            \begin{minipage}[t]{0.45\linewidth}
            \centering
            \includegraphics[height=1.7in,width=\textwidth]{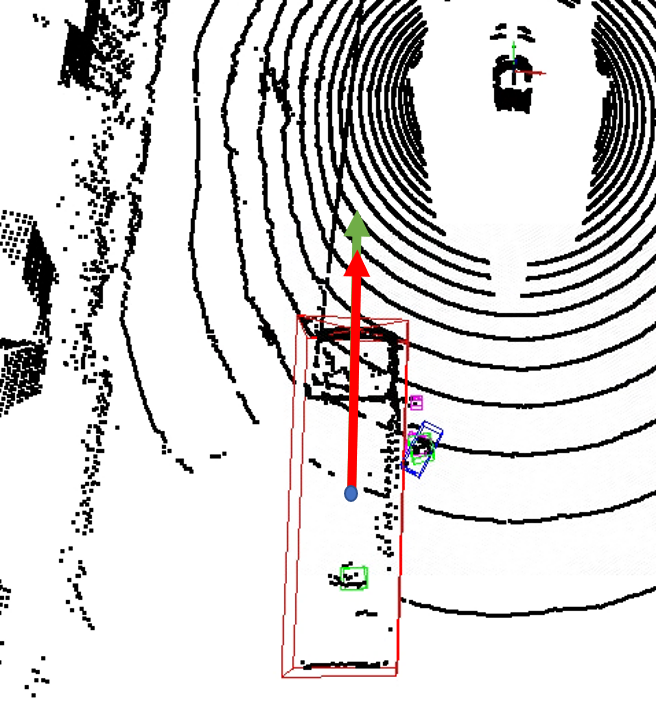} \\
            \end{minipage}
        }
        
        \caption{Comparison of the original 3D detections and the refined 3D detections. The green arrows indicate the orientations of ground truth while the red arrows represent the orientations of the original detection and the refined bounding boxes.}
        \vspace{-0.2cm}
    \end{figure}

    \textbf{KITTI:} KITTI consists of 7481 training frames and 7518 testing frames with 3D and 2D annotations of cars, pedestrians, and cyclists on the streets. Each class is further divided into three levels: easy, moderate, and hard according to their object size, truncation level, and occlusion level. Different methods are evaluated using Average Precision (AP) calculated by 3D IoU. In this paper, we split the original training samples into 3740 training samples used for the learning-based method and 3741 validation samples.

    \subsection{Benchmark results}
    \textbf{LiDAR-based baseline:} To evaluate the improvements on LiDAR-based 3D detectors, we apply our proposed method for combinations of 2D detector YOLOv5 \cite{b23} and different 3D detectors such as PointPillars \cite{b25}, SECOND \cite{b16}, and CenterPoint \cite{b27}. All 3D detection models are trained with 8 GTX 1080Ti GPUs and can be downloaded from \cite{b28}. We set the original LiDAR-based 3D detector as the baseline and the performances of our method with each corresponding baseline on nuScenes trainval dataset are summarized in Table \uppercase\expandafter{\romannumeral1}. As expected, GOOD performs better and shows a 6.2\%, 5.0\%, and 4.2\% relative increase in NDS compared with PointPillars, SECOND, and CenterPoint respectively while the performances on mAP score are even better with 9.1\%,  6.4\%, and 3.9\% relative increase respectively. As for the error index, although GOOD performs slightly worse than SECOND on mAOE score (around 1.4\%) since the 3D cuboids reconstructed by the $add$ module may have the wrong orientations with constraints from only a single frame, it still has better performances on comparisons with all other baseline methods and error indexes.

    \textbf{Multi-modal-based baseline:} We compare our method with the learning-based late fusion strategy, CLOCs \cite{b14}, by leveraging SECOND as the 3D detector on KITTI dataset for the sake of fairness which is the only supported 3D detector and dataset on the current open source version of CLOCs. Additionally, considering that KITTI is not sampled sequentially so that it is not available for tracking, GOOD adopted here does not contain the 3D MOT pipeline. The IoU threshold for car, pedestrian, and cyclist are set as 0.7, 0.5, and 0.5 separately. The results are summarized in Table \uppercase\expandafter{\romannumeral2}, which indicates that our proposed method performs well for car and pedestrian and even better than CLOCs at easy level with nearly 2\% improvement on 3D AP. Since occlusion and truncation are not severe, the optimization constraints brought by 2D detection are feasible and effective in this situation. Since the shapes of different cyclists may vary frequently at moderate or hard levels and the number of cyclists on the validation set is relatively modest in total, there are relatively less improvements compared with CLOCs due to the introduction of shape residual or several false cases. However, it is worth mentioning that the performance of GOOD can be further improved with the aid of 3D MOT as evaluated in the following ablation study experiments.

    \subsection{Ablation study}
    \textbf{Optimization and 3D MOT pipeline:} We evaluate the effectiveness of each component of our proposed pipeline, including $keep$ $or$ $delete$ (for short KoD), $refine$, $add$, and 3D MOT. The results are shown in Table \uppercase\expandafter{\romannumeral3}, where we can conclude that the KoD module can help improve mAP score by removing the unmatched 3D instances which have low detection scores since they may be the false or redundant detections. The $refine$ module contributes the most in reducing errors since the constraints from 2D detection can help correct the state of 3D detection as shown in Fig. 4, from which we can see that the red bounding box representing a truck in original 3D detections is slightly askew while its orientation can be adjusted under the constraint from 2D detection bounding box. Besides, a significant improvement on mAP score can be observed with the aid of the $add$ module and Fig. 3 provides an example to illustrate our proposed depth and z-coordinate of the reconstructed cuboid adjustment algorithm vividly. The introduction of 3D MOT is also beneficial to both the mAP score and the error indexes.

    \textbf{Matching pipeline:} In addition, we also design a controlled experiment to demonstrate the effectiveness of our matching method in this paper as shown in Table \uppercase\expandafter{\romannumeral4}. The result shows that compared with the one-sided nearest-neighbor matching, mutual-sided nearest-neighbor matching behaves better on NDS and mAP score.

    \begin{table}[htbp] 
        \caption{\label{tab:test}Performance comparison of different matching methods. We choose PointPillars as the 3D detector and YOLOv5 as the 2D detector.}
        \centering
        \vspace{0.3cm}
        \setlength{\tabcolsep}{1.8mm}
        \begin{threeparttable}
            \begin{tabular}{ccc} 
                \toprule 
                Method & NDS $\uparrow$ & mAP $\uparrow$ \\ 
                \midrule 
                Ours (one-sided nearest-neighbor \cite{b1}) & 0.716 & 0.639 \\
                Ours (mutual-sided nearest-neighbor)& \textbf{0.721} & \textbf{0.648} \\
                \bottomrule 
            \end{tabular} 
        \end{threeparttable}
    \end{table}

\section{Conclusions}

This paper proposes a general optimization-based fusion framework for 3D object detection (GOOD) that leverages the output candidates from LiDAR and camera and is available for any combination of 3D/2D detectors to improve the accuracy and robustness of 3D detection. To testify to our proposed method, experiments on nuScenes and KITTI dataset are conducted, and the final results show that our proposed method outperforms the the original 3D detection baseline by an appreciable margin on nuScenes dataset and achieves competitive performance with the learning-based late fusion CLOCs on KITTI dataset. To the best of our knowledge, this is the first optimization-based late fusion framework for multi-modal 3D object detection without retraining additional models when the scene is switched.

There are several directions for future work. Since the $add$ module only utilizes a single frame from the camera, which lead to inaccurate state estimation of the 3D bounding box through the introduction of depth and z-coordinate adjustment algorithm, 3D cuboid reconstruction aided by multiple frames is considered. Apart from this, more experiments on other datasets are also feasible to validate the effectiveness of our method.







\bibliography{dense}

\end{document}